\documentclass[letterpaper, 10 pt, conference]{ieeeconf}  

\IEEEoverridecommandlockouts                              

\usepackage{graphicx}
\usepackage{color}
\usepackage{cite}
\usepackage{hyperref}
\hypersetup{
hidelinks,
colorlinks=true,
linkcolor=red,
citecolor=green,
filecolor=magenta,
urlcolor=blue
}
\usepackage{multirow}
\usepackage{enumerate}
\usepackage{bm}
\usepackage{booktabs}
\usepackage{blindtext}
\usepackage[T1]{fontenc}
\usepackage{mathptmx} 
\DeclareMathAlphabet{\mathcal}{OMS}{cmsy}{m}{n}
\usepackage{times} 
\usepackage{amsmath} 
\usepackage{amssymb}  
\usepackage[ruled,linesnumbered]{algorithm2e}

\usepackage{xcolor}

\usepackage{bbm}
\usepackage{booktabs}
\usepackage{upgreek}
\usepackage{mathtools,xparse}
\usepackage{colortbl}

\newcommand{\degree}{^\circ}

\newcommand{\PreserveBackslash}[1]{\let\temp=\\#1\let\\=\temp}
\newcolumntype{C}[1]{>{\PreserveBackslash\centering}p{#1}}
\newcolumntype{R}[1]{>{\PreserveBackslash\raggedleft}p{#1}}
\newcolumntype{L}[1]{>{\PreserveBackslash\raggedright}p{#1}}

\DeclarePairedDelimiter{\norm}{\lVert}{\rVert}
\NewDocumentCommand{\normL}{ s O{} m }{%
  \IfBooleanTF{#1}{\norm*{#3}}{\norm[#2]{#3}}_{L_2($\Omega$)}%
}

\DeclareFontFamily{U} {cmmi}{}

\DeclareFontShape{U}{cmmi}{m}{n}{
  <-6> cmmi5
  <6-7> cmmi6
  <7-8> cmmi7
  <8-9> cmmi8
  <9-10> cmmi9
  <10-12> cmmi10
  <12-> cmmi12}{}

\DeclareSymbolFont{Xcmmi} {U} {cmmi}{m}{n}

\DeclareMathSymbol{\psi}{\mathord}{Xcmmi}{32}

\title{DiGNet: Learning Scalable Self-Driving Policies for Generic Traffic Scenarios with Graph Neural Networks}

\author{Peide~Cai$^1$,
        Hengli~Wang$^1$,
        Yuxiang~Sun$^2$,
        and~Ming~Liu$^1$,~\IEEEmembership{Senior Member,~IEEE}
\thanks{This work was supported by Zhongshan Municipal Science and Technology Bureau Fund, under project ZSST21EG06, Collaborative Research Fund by Research Grants Council Hong Kong, under Project No. C4063-18G, and Department of Science and Technology of Guangdong Province Fund, under Project No. GDST20EG54, awarded to Prof. Ming Liu. \textit{(Corresponding author: Ming Liu.)}}

\thanks{$^1$ECE, the HKUST; $^2$ME, PolyU of HK. (email: pcaiaa@connect.ust.hk, hwangdf@connect.ust.hk, yx.sun@polyu.edu.hk, eelium@ust.hk).}

}
\IEEEaftertitletext{\vspace{-0.6\baselineskip}}
\begin{document}

\maketitle
\thispagestyle{empty}
\pagestyle{empty}

\begin{abstract}
Traditional decision and planning frameworks for self-driving vehicles (SDVs) scale poorly in new scenarios, thus they require tedious hand-tuning of rules and parameters to maintain acceptable performance in all foreseeable cases. Recently, self-driving methods based on deep learning have shown promising results with better generalization capability but less hand engineering effort. However, most of the previous learning-based methods are trained and evaluated in limited driving scenarios with scattered tasks, such as lane-following, autonomous braking, and conditional driving. In this paper, we propose a graph-based deep network to achieve scalable self-driving that can handle massive traffic scenarios. Specifically, more than 7,000 km of evaluation is conducted in a high-fidelity driving simulator, in which our method can obey the traffic rules and safely navigate the vehicle in a large variety of urban, rural, and highway environments, including unprotected left turns, narrow roads, roundabouts, and pedestrian-rich intersections. Demonstration videos are available at \url{https://caipeide.github.io/dignet/}.

\end{abstract}

\thispagestyle{empty}
\pagestyle{empty}

\section{Introduction}

Self-driving in complex and dynamic traffic scenarios are quite challenging. For one thing, there are diverse road topologies and scenarios (e.g., roundabouts, multi-lane streets and intersections) to consider\cite{Chen2019DeepIL}. For another, the complex and coupled interaction among multiple road agents (e.g., pedestrians, bicyclists and vehicles) are hard to model\cite{Chen2020RobotNI}. In addition, various traffic rules should be obeyed, such as traffic light, lane markings and speed limit. 

Classical decision and planning methods are largely rule-based, and are difficult to generalize to new scenarios. Therefore, they require time-intensive and iterative hand tuning of rules and parameters to maintain performance\cite{Zeng2019EndToEndIN}. In recent years, as an alternative way, deep learning has advanced the SDV technology to a great extent. The ability to \textit{learn and self-optimize} its behavior from data alleviates the laborious engineering maintenance for modeling all foreseeable scenarios, making a deep driving model well suited to self-driving problems in high-dimensional, nonlinear and dynamic environments\cite{Cai2020ProbabilisticEV, Chen2019LearningBC,OhnBar2020LearningSD, codevilla2018end, codevilla2019exploring, Tai2019VisualbasedAD}. However, current learning-based models have not been well designed for scalable self-driving in a uniform setup. Specifically, most works only focus on scattered tasks such as conditional urban driving\cite{OhnBar2020LearningSD, codevilla2018end, codevilla2019exploring,Chen2019LearningBC,Tai2019VisualbasedAD}, lane-following\cite{Xu2017EndtoEndLO}, lane-changing\cite{Wang2018ARL} or collision avoidance\cite{Porav2018ImminentCM, Woo2020CollisionAF}. In short, high-performance scalable driving models that can handle massive traffic scenarios still seem out of reach. In order to solve this problem, in this paper we propose a graph-based deep neural network \textit{DiGNet} (driving in graphs) for scalable self-driving. Specifically, we use graph attention networks (GAT)\cite{velickovic2018graph} to model the complex interactions among traffic agents, use semantic bird's-eye view (BEV) images to model road topologies and traffic rules, and adopt the variational auto-encoder (VAE)\cite{kingma2014autoencoding} to extract compact yet effective environmental features. 

The main contribution of this paper are listed as follows: (1) We propose a specially designed network architecture for self-driving in generic traffic scenarios. It considers both driving contexts and the interaction among dynamic road agents. (2) We demonstrate the scalable and safe self-driving performance of the proposed method through extensive closed-loop evaluation in both seen and unseen urban/rural/highway environments.
    
\begin{figure*}[t]
        \centering
        \includegraphics[width = 2\columnwidth]{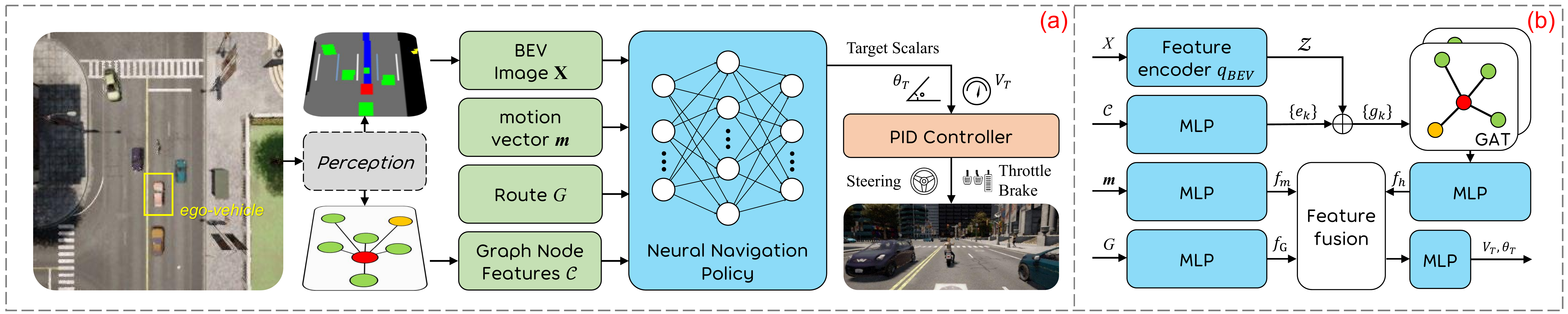}
        \caption{(a) Schematic overview of the proposed method for self-driving. (b) The data flow inside the neural navigation network (DiGNet). In this work, we assume the input information is accessible with a functioning perception system or with a vehicle-to-everything (V2X) module, and focus on \textit{learning to act} of SDVs.}
        \label{fig:architecture}
        \vspace{-0.35cm}
\end{figure*}

\section{Related Work}

\textbf{End-to-end self-driving.} With the powerful representation capabilities of deep neural networks, \textit{end-to-end} driving approaches \cite{Xu2017EndtoEndLO, codevilla2018end, Chen2019LearningBC, Tai2019VisualbasedAD, Cai2020ProbabilisticEV, OhnBar2020LearningSD, codevilla2019exploring} directly take as input the raw sensor readings (e.g., LiDAR point clouds, camera images) to output control commands such as steering angle and throttle. These methods are mostly trained with imitation learning where the agent tries to mimic an expert driver. For example, based on large amount of labelled data, Xu \textit{et al.} \cite{Xu2017EndtoEndLO} developed an end-to-end architecture to predict future ego-motions. However, it only targets lane-following tasks and lacks closed-loop experiments. Codevilla \textit{et al.} \cite{codevilla2018end} proposed a conditional imitation learning approach that splits the network into multiple branches for discrete tasks such as \textit{follow lane} and \textit{turn left/right}. Follow-up works include \cite{Chen2019LearningBC}, \cite{codevilla2019exploring} and \cite{Tai2019VisualbasedAD}. However, these methods cannot handle complex road topologies such as multi-lane streets or roundabouts. Recently, Cai \textit{et al.}\cite{Cai2020ProbabilisticEV} used global routes as direction to achieve robust end-to-end navigation in complex dynamic environments with multi-modal sensor fusion. The drawback of this work is the neglect of both traffic rules and efficient interaction with other road agents (e.g., bypassing a vehicle blocked ahead). To summarize, it is quite challenging to learn a direct mapping from high dimensional sensory observations to low dimensional control output, as the end-to-end approaches conflates two aspects of driving: \textit{learning to see} and \textit{learning to act}. Therefore, they suffer from the domain gap problem in the perception stage, which prevents them from scalable self-driving (demonstrated in Sec. \ref{sec:experiments}). 

\textbf{Learning to drive by semantic abstraction.} Recently, there arises another framework that uses \textit{semantic} perception results (e.g., HD maps\cite{Chen2019DeepIL}, birdeye views (BEVs)\cite{bansal2019chauffeurnet}, and occupancy maps\cite{sadat2020perceive}) to learn driving policies. Compared with the redundant sensory observations of end-to-end driving methods, these semantic input representations are both concise and informative, and have better environmental consistency, making the training process focus on \textit{learning to act}. For example, the policy network of \cite{sadat2020perceive} takes as input hybrid features composed of road maps, traffic light, route plans and dynamic objects to produce waypoints to follow. This has the advantage to help the network learn meaningful contextual cues behind the human driver's action and achieve more complex driving behaviors. However, \cite{sadat2020perceive} mainly shows its results on offline \textit{logged data}, and only performs closed-loop evaluation in simple environments with at most 2 obstacles. This problem also exists in other similar works\cite{bansal2019chauffeurnet}. According to \cite{Codevilla2018OnOE}, the driving performance can vary a lot between offline open-loop tests and online closed-loop tests, and only the latter can reveal the actual performance. In this work, our input representation is similar to that of \cite{bansal2019chauffeurnet, sadat2020perceive} but differently, we focus more on dynamic, interactive and large-scale \textit{closed-loop} driving performance. 

\textbf{Graph neural networks (GNNs).} Many real-world problems can be modeled with graphs where the nodes contain features of different entities, and edges represent interactions between entities. A challenge in learning on graphs is to find an effective way to get a meaningful aggregated feature representation to facilitate downstream tasks. Recently, graph convolutinal networks (GCNs) \cite{Defferrard2016ConvolutionalNN} have shown to be effective in many applications such as social networks, personalized recommendation and link prediction. GCN generalizes the 2D convolution on grids to graph-structured data. When training a GCN, a fixed adjacency matrix is commonly adopted to aggregate feature information of neighboring nodes. Graph attention network (GAT)\cite{velickovic2018graph} is a GCN variant, and it aggregates node information with weights learned in a self-attention mechanism. Such adaptiveness of GAT makes it more effective than GCN in graph representation learning.


\section{Formulation}

Fig. \ref{fig:architecture} shows the structure of the proposed DiGNet for scalable self-driving. The goal is to drive safely in a large variety of complex and dynamic outdoor environments for point-to-point navigation. To this end, we use GAT to model the complicated interactions among road agents, and BEV images to represent various road structures. In addition, we choose the open-source CARLA simulation (0.9.9)\cite{dosovitskiy2017carla} to train and evaluate our system, since it possesses abundant vehicle models and maps close to the real world.

\subsection{Driving Scene Representation}
\begin{figure}[t]
        \centering
        \includegraphics[width = \columnwidth]{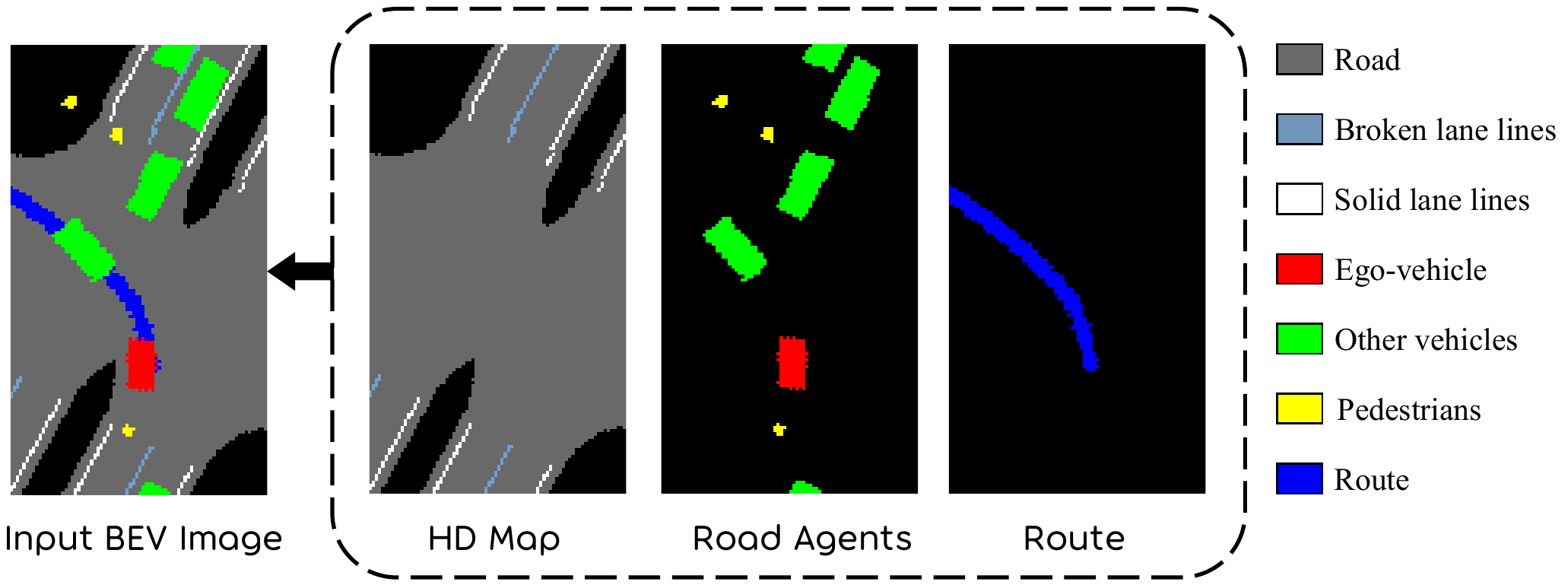}
        \caption{BEV representation as one of the input to our network. It includes information of HD map, road agents, routes and traffic light.}
        \label{fig:bev_masks}
\end{figure}

We rasterize different semantic elements (e.g., lane marking, obstacles) into multiple binary channels to form a concise and informative scene representation. As shown in Fig. \ref{fig:bev_masks}, our BEV input is composed of the following three parts with seven channels in total: (1) \textbf{High-definition (HD) map}: The HD map contains the drivable area and lane markings (solid and broken lines). Leveraging map information to learn driving policies is very helpful because it provides valuable structural priors on the motion of surrounding road agents. For example, vehicles normally drive on lanes rather than on sidewalks. Another benefit is that vehicles need to drive according to traffic rules, such as not crossing solid lane markings. (2) \textbf{Routes and traffic light}: The route is provided by a global planner implemented with $A^*$. The route channel will be set to blank when the traffic light that affects the ego-vehicle turns to red, otherwise it maintains its normal value. (3) \textbf{Road agents}: We render the ego-vehicle, other vehicles and pedestrians in three other channels.

In this work, our region of interest is $W$=20 m wide (10 m to each side of the ego-vehicle) and $H$=35 m long (25 m front and 10 m behind of the ego-vehicle). The image resolution is set to 0.25 m/pixel, which finally results in a binary BEV input $\mathbf{X}$ of size 80$\times$140$\times$7.

\subsection{Data Collection}
\label{subsec:data-collection}
We collect 260 expert driving episodes in CARLA for training. The collected dataset lasts about 7.6 hours and covers a driving distance of 150 km. For each data collection episode, we set random routes ranging from 300 m to 1500 m in four maps of CARLA: \texttt{Town03}, \texttt{Town05}, \texttt{Town06} and \texttt{Town07}, covering urban, rural and highway scenarios. We also set roaming pedestrians and vehicles, which are controlled by the AI engine of CARLA, to form realistic traffic scenarios.

\section{Methodology}

\subsection{Learning the Context Embedding}
\label{subsec:vae}

\begin{figure}[t]
        \centering
        \includegraphics[width = \columnwidth]{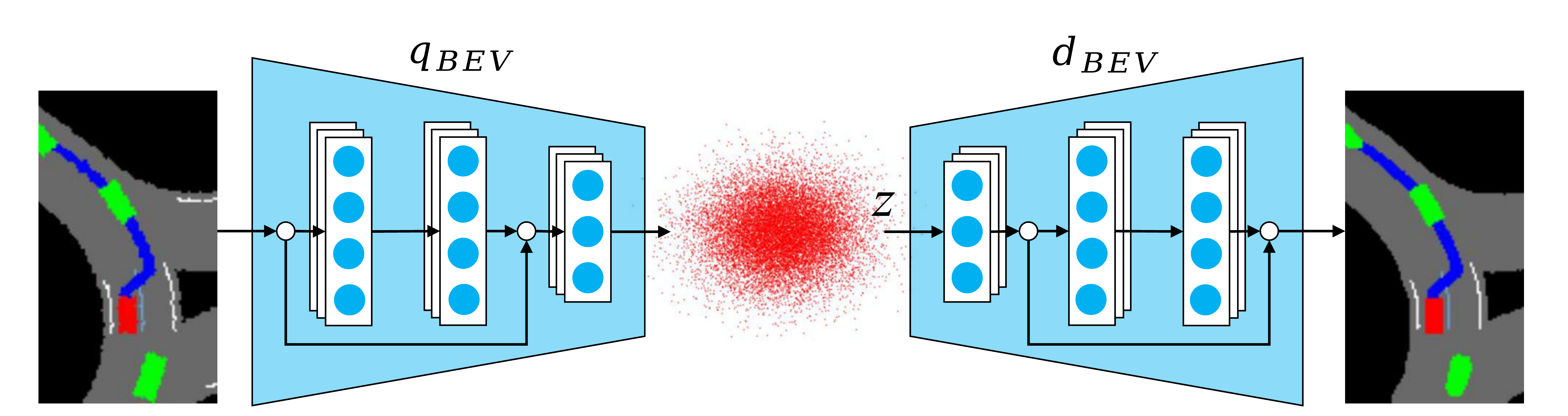}
        \caption{VAE architecture. The input image is encoded into a latent space, from which the sampled vector $\bm{z}$ can be decoded back into a reconstructed image similar to the input.}
        \label{fig:vae_net}
\end{figure}

Although we replace the complex raw sensor data with more informative and concise BEV semantic masks, it is still quite high-dimensional compared to the vehicle state variables (e.g., speed, location, etc.). Such high-dimensionality makes it not only hard to learn good policies in data scarce tasks, but also suffer from over-fitting problems. Therefore, a lower-dimensional embedding for the multi-channel BEV input is needed for us to train a high-performance driving policy. To this end, we first train a variational autoencoder (VAE) on the collected BEV images from Section \ref{subsec:data-collection}. As shown in Fig. \ref{fig:vae_net}, this method can help to summarize the key geometrical properties of environments in a low-dimensional latent vector $\bm{z}$. More specifically, we adopt the $\beta$-VAE\cite{Higgins2017betaVAELB} and minimize the variational lower bound with encoder $q_{BEV}$ and decoder $d_{BEV}$:
\begin{equation}
\mathcal{L}_\text{VAE}=\beta\cdot \text{D}_\text{KL}\left(q_{BEV}(\bm{z} \mid \mathbf{X}) \| p(\bm{z})\right) + \left\|d_{BEV}(\bm{z})-\mathbf{X}\right\|_{2}^{2},
\label{eq:vae}
\end{equation}
where $\text{D}_\text{KL}(\cdot)$ is the Kullback-Leibler (KL) divergence. The encoder $q_{BEV}(\bm{z} \mid \mathbf{X})$ takes as input the raw image and returns the mean $\mu$ and variance $\sigma^2$ of a normal distribution, such that $\bm{z} \sim \mathcal{N}\left(\mu, \sigma^{2}\right)$. $p(\bm{z})$ is the prior on the latent space, modeled as the standard normal distribution $\mathcal{N}\left(0, I\right)$. The second term of (\ref{eq:vae}) is the MSE loss between the raw and reconstructed images ($\mathbf{X}, d_{BEV}(\bm{z})$). The parameter $\beta$ provides a trade-off between these two types of losses. 

In this work, we set $\beta=0.01$ and $\bm{z}\in \mathbb{R}^{512}$. We implement the VAE based on the ResNet18 architecture, where the decoder uses a combination of upsampling and convolutions for image reconstruction. Note the parameters of VAE are fixed during later policy training for stable performance. 


\subsection{Graph Modeling of Driving Scenes}
\label{subsec:GAT}
\subsubsection{Network Architecture}


As shown in Fig. \ref{fig:architecture}, we use GAT, which is composed of multiple graph layers, to model the interaction among road agents during driving. The input to the i-th layer is a set of node features, $\{\bm{h}^i_1,  \bm{h}^i_2, ...,  \bm{h}^i_N\}$, $ \bm{h}^i_k \in \mathbb{R}^{F^i}$, where $N$ is the number of nodes (agents, including the ego-vehicle), and $F^i$ is the dimension of features in each node. Then, information of each node $k$ are propagated to the neighboring nodes $\mathcal{N}_k$ and being used to update the node features via a self-attention mechanism, which produces the output of the layer:
\begin{equation}
\bm{h}^{i+1}_{k}=\sigma(\sum_{j \in \mathcal{N}_{i}} \alpha_{kj}(\bm{h}_{k}^{i},\ \bm{h}_{j}^{i}) \mathbf{W} \bm{h}_{j}^{i}),
\label{eq:graoh attention}
\end{equation}
where $\sigma(\cdot)$ is the nonlinear activation function, e.g., ReLU; $\mathbf{W} \in \mathbb{R}^{F^{i+1}\times F^{i}}$ is a shared weight matrix to be applied to each node for expressive feature transformation; $\alpha_{kj}(\cdot,\cdot)$ means the importance of node $j$ to node $k$, and it is the normalized attention coefficients computed with shared weight vector $\vec{\mathbf{a}} \in \mathbb{R}^{2 F^{i+1}}$:
\begin{equation}
\alpha_{kj} (\bm{h}_{k}^{i},\ \bm{h}_{j}^{i} ) = \frac{\text{exp} ( \sigma ( \vec{\mathbf{a}}^T [\mathbf{W}\bm h^i_k||\mathbf{W}\bm h^i_j  ] ) )}{\sum_{m\in \mathcal{S}_k} (\sigma (\vec{\mathbf{a}}^T [\mathbf{W} \bm h_k^i || \mathbf{W} \bm h_m^i   ]  ) )},
\end{equation}
where $||$ represents the concatenation operation. Furthermore, We follow the \textit{multi-head attention} method in \cite{velickovic2018graph} to stabilize the learning process. Specifically, $S^i$ independent graph networks execute the transformation of (\ref{eq:graoh attention}) and their features are concatenated to produce the output of the i-th layer:
\begin{equation}
\bm{h}_{k}^{i+1}=\mathop{\|}\limits_{s=1}^{S^i} \sigma(\sum_{j \in \mathcal{N}_{i}} \alpha_{kj}^s(\bm{h}_{k}^{i},\ \bm{h}_{j}^{i}) \mathbf{W}^s \bm{h}_{j}^{i}).
\end{equation}

For the final layer, we employ \textit{averaging} among multiple heads rather than concatenation.

\subsubsection{Implementation Details}
In this work, we adopt a two-layer GAT and set $S^1, S^2=5, \ F^1,F^2=256$. The input features $\mathcal{C}$ include motion state information for each node (road agent) in the ego-vehicle's local coordinate. For node $k \in \{1,2, ... ,N\}$, the input feature is a 10-dimensional vector:
\begin{equation}
    \bm{c}_k=\left\{x, y, d, \psi, vx, vy, ax, ay, w, l\right\},
\end{equation}
which includes its location ($x, y$), distance to the ego-vehicle ($d$), yaw angle ($\psi$), velocity ($vx, vy$), acceleration ($ax, ay$) and size (width $w$ and length $l$). In the spirit of \cite{Chen2020RobotNI}, we first pass each node state $\bm{c}_k \in \mathcal{C}$ through a multilayer perceptron (MLP) to produce a feature vector $\bm{e}_k \in \mathbb{R}^{128}$ for sufficient expressive power. For context-aware graph modeling, we then concatenate $\bm{e}_k$ with $\bm{z}$ derived from the VAE introduced in Section \ref{subsec:vae}, to generate the mixed vector $\bm{g}_k$. Then the set $\{\bm{g}_k\}$ are sent to GAT to output the final aggregated feature $\bm{h}^o_k  \in \mathbb{R}^{256}$, which represents the internal interactions on each node $k$. We are interested in the result $\bm{h}^o_1$ of the first node, which represents the influence on the ego-vehicle.

\subsection{Task-relevant Feature Embedding}
\label{subsec:task-relevant}
In addition to the interaction feature $\bm{h}^o_1$ derived from the above section, we also need task-relevant information of the ego-vehicle to achieve mannered goal-directed self-driving, and it is composed of two other feature vectors. The first vector $\bm{m} \in \mathbb{R}^{13}$ describes the ego-motion as follows:
\begin{equation}
    \bm{m} = \left\{ \tau_s, \tau_t , \tau_b , v_{lim}, vx, vy, ax, ay, e_v, e_{cte}, e_{heading},   F_{l}, F_{r}\right\},
\label{eq:ego-motion-vector}
\end{equation}
where $\tau_s, \tau_t \text{ and } \tau_b$ indicate the current control command of the ego-vehicle: steering angle, throttle and brake. $v_{lim}$ means the traffic speed limit, $e_v$ means the difference between $v_{lim}$ and current vehicle speed, $e_{cte}$ is the cross track error between the vehicle location and the reference route, and $e_{heading}$ is the heading angle error calculated with vector field guidance\cite{Nelson2007VectorFP}. $F_{l(r)}$ is a binary indicator of the lane marking on the left (right) side of the ego-vehice. The indicator is set to 1 if the lane marking is crossable (e.g., a broken line), and set to 0 otherwise (e.g, a solid line).

The second vector is a route vector $G\in \mathbb{R}^{150}$:
\begin{equation}
G=\left\{\left(x_{k}, y_{k}\right) \mid 1 \leq k \leq 75\right\} \subset G_{f},
\label{eq:local-relevant-route}
\end{equation}
where $G_f$ is the full high-level route from the start point to the destination. During navigation, we down-sample $G_f$ into local relevant route $G$ based on the ego-vehicle's location. Specifically, the first waypoint $(x_1,y_1)$ in $G$ is the closest waypoint in $G_f$ to the current vehicle location, and the distance of every two adjacent points is 0.4 m. Note the redundancy of route information in $G$ and BEV is meaningful, where the state vector here provides more specific waypoint locations for the vehicle to follow, while the route mask of BEV can indicate if there are any obstacles on conflicting lanes of planned routes, as well as encode traffic light information.

Finally, we adopt three MLPs to transform $\bm{m}$, $G$ and $\bm{h}^o_1$ separately into feature vectors $f_m$, $f_G$, $f_h \in \mathbb{R}^{1024}$ for further processing.

\begin{table*}[t]
\newcommand{\tabincell}[2]{\begin{tabular}{@{}#1@{}}#2\end{tabular}}
\newcommand{\NA}{---}
        \setlength{\abovecaptionskip}{-1pt}
        \renewcommand{\arraystretch}{1.3}
        \definecolor{minigray}{rgb}{0.92, 0.92, 0.92}
        \caption{Closed-loop Evaluation Results of Different Models on Six Maps in CARLA. SR Means the Success Rate (\%), and DS Means the Driving Score. Larger Numbers Are Better. The Bold Font Highlights the Best Results in Each Column.}
        \label{tab:evaluation}
        \centering
        \begin{tabular}{l C{0.8cm} C{0.8cm} C{0.8cm} C{0.8cm} C{0.8cm} C{0.8cm} C{0.8cm} C{0.8cm} C{0.8cm} C{0.8cm} C{0.8cm} C{0.8cm} }
        
        \toprule
        {}&
        \multicolumn{8}{c}{{Training Towns}}&
        \multicolumn{4}{c}{{Unseen Towns}} \\
        \cmidrule(lr){2-9}\cmidrule(lr){10-13}
        \multirow{2}{*}{}&
        \multicolumn{2}{c}{\texttt{Town03}} &
        \multicolumn{2}{c}{\texttt{Town05}} & 
        \multicolumn{2}{c}{\texttt{Town06}} & 
        \multicolumn{2}{c}{\texttt{Town07}} &
        \multicolumn{2}{c}{\texttt{Town04}} &
        \multicolumn{2}{c}{\texttt{Town10}}\\
        \multirow{2}{*}{}& \multicolumn{2}{c}{\textit{urban}} & \multicolumn{2}{c}{\textit{urban}}
        & \multicolumn{2}{c}{\textit{highway}} & \multicolumn{2}{c}{\textit{rural}}
        & \multicolumn{2}{c}{\textit{mixed}} & \multicolumn{2}{c}{\textit{urban}}
        \\
        \midrule
        Models & SR & DS & SR & DS & SR & DS & SR & DS & SR & DS & SR & DS\\
        \midrule
        
        \texttt{CILRS} & 43.2 & 0.56 & 38.2 & 0.52 & 35.2 & 0.48 & 44.8 & 0.59 & 16.2 & 0.28 & 1.2 & 0.11 \\
        \texttt{MLP} & 62.3 & 0.72 & 60.0 & 0.69 & 72.2 & 0.79 & 75.0 & 0.82 & 60.5 & 0.70 & 38.2 & 0.53 \\
        \texttt{GCN(U)} & 68.8 & 0.77 & 70.2 & 0.78 & 73.8 & 0.80 & 73.0 & 0.80 & 68.5 & 0.76 & 55.2 & 0.66 \\
        \texttt{GCN(D)} & 74.2 & 0.80 & 71.5 & 0.79 & 74.8 & 0.81 & 75.2 & 0.82 & 64.0 & 0.72 & 62.5 & 0.71 \\
        \texttt{GAT} & 76.0 & 0.82 & \textbf{81.5} & \textbf{0.87} & \textbf{78.2} & \textbf{0.83} & 76.8 & 0.83 & 71.0 & 0.78 & \textbf{68.0} & \textbf{0.75} \\
        \texttt{DiGNet(CTL)} & 72.8 & 0.79 & 69.8 & 0.77 & 73.2 & 0.79 & 73.5 & 0.80 & 67.5 & 0.75 & 49.0 & 0.58 \\
        \rowcolor{minigray}
        \texttt{DiGNet}  (\textit{ours})& \textbf{80.2} & \textbf{0.85} & 75.0 & 0.82 & \textbf{78.2} & \textbf{0.83} & \textbf{80.0} & \textbf{0.85} &\textbf{72.8} & \textbf{0.80} & 67.2 & \textbf{0.75} \\
        
        \bottomrule
        \end{tabular}
        \vspace{-0.2cm}
\end{table*}

\subsection{Driving Policy Training}
\label{subsec:training}

With components defined above, the derived feature vectors $f_m$, $f_G$, $f_h$ are concatenated and then processed with a MLP to produce the final output for navigating the vehicle. Different from most of the previous end-to-end driving methods\cite{Chen2019LearningBC,OhnBar2020LearningSD, codevilla2018end, codevilla2019exploring, Tai2019VisualbasedAD} that directly regress vehicle control commands (e.g., steering and throttle), we adopt a mid-level output representation indicating the target speed $v_{T}$ and course angle $\theta_{T}$. For implementation, to bound the range of $v_{T}$ and $\theta_{T}$, we regress two scalars of target speed ($\kappa_v \in [0,1]$) and course angle ($\kappa_{\theta} \in [-1,1]$). Then the target control values can be computed as follows:
\begin{equation}
    v_{T} = v_{lim} \times \kappa_v,\ \  \theta_T = 90\degree \times \kappa_{\theta}.
\end{equation}
These mid-level indicators are finally translated by a PID controller to generate the device-level control commands, i.e., steering, throttle and brake. In this way, we can reduce the burden of control and inject the information of speed limit to the model output, which leads to more flexible policy models. We use L1 loss in terms of $v_T$ and $\theta_T$ to train the policy by imitation learning.

\section{Experiments and Discussion}
\label{sec:experiments}
\subsection{Training setup}
We train our network \texttt{DiGNet} on the collected dataset introduced in Section \ref{subsec:data-collection}. The split ratio of training, validation and testing set is set to 17:1:2, leading to 240K training samples. We set the batch size to 256, and use the Adam optimizer with a learning rate of 0.0001. For comparison, we also train six other baselines, which can be divided into the following two categories:

1) \textit{End-to-end} self-driving method.

\begin{itemize}
    \item \texttt{CILRS}. It is one of the state-of-the-art vision-based end-to-end driving method\cite{codevilla2019exploring}, which takes as input the front-facing camera images and vehicle speeds to output control commands. This baseline is added to verify that policies trained with abstracted semantics have better generalization capability.
\end{itemize}

2) \textit{Graph state}-based methods. These networks are named based on how they process the node features $\mathcal{C}$.

\begin{itemize}

    \item \texttt{GCN(U)}. It takes as input $\{\mathcal{C}, G, \bm{m}\}$ to produce the output. It adopts a two-layer GCN to process the node features $\mathcal{C}$. We refer to the baseline \textit{U-GCNRL} introduced in \cite{Chen2020RobotNI} and set the adjacency matrix of GCN with \underline{u}niform weihgts.

    \item \texttt{GCN(D)}. It is similar to \texttt{GCN(U)} but uses \underline{d}istance-related weights in its adjacency matrix. It adopts a straightforward intuition that obstacles closer to the ego-vehicle should exert a stronger influence. This network follows the idea of \textit{D-GCNRL} introduced in \cite{Chen2020RobotNI}. 
    
    \item \texttt{GAT}. Its difference to our proposed method is the non-use of the context embedding $\bm{z}$. 
    
    \item \texttt{DiGNet(CTL)}. It is a variant of our method utilizing both graph state and context embedding $\bm{z}$. However, it directly generates device-level control commands (i.e., steering, throttle and brake), rather than the mid-level target speed/angle scalars.

    \item \texttt{MLP}. It uses MLP rather than GNNs to process the node features. This baseline is added to verify the advantage of GNN to handle interactions among different nodes.
    
\end{itemize}

\begin{figure*}[t]
        \centering
        \includegraphics[width = 2\columnwidth]{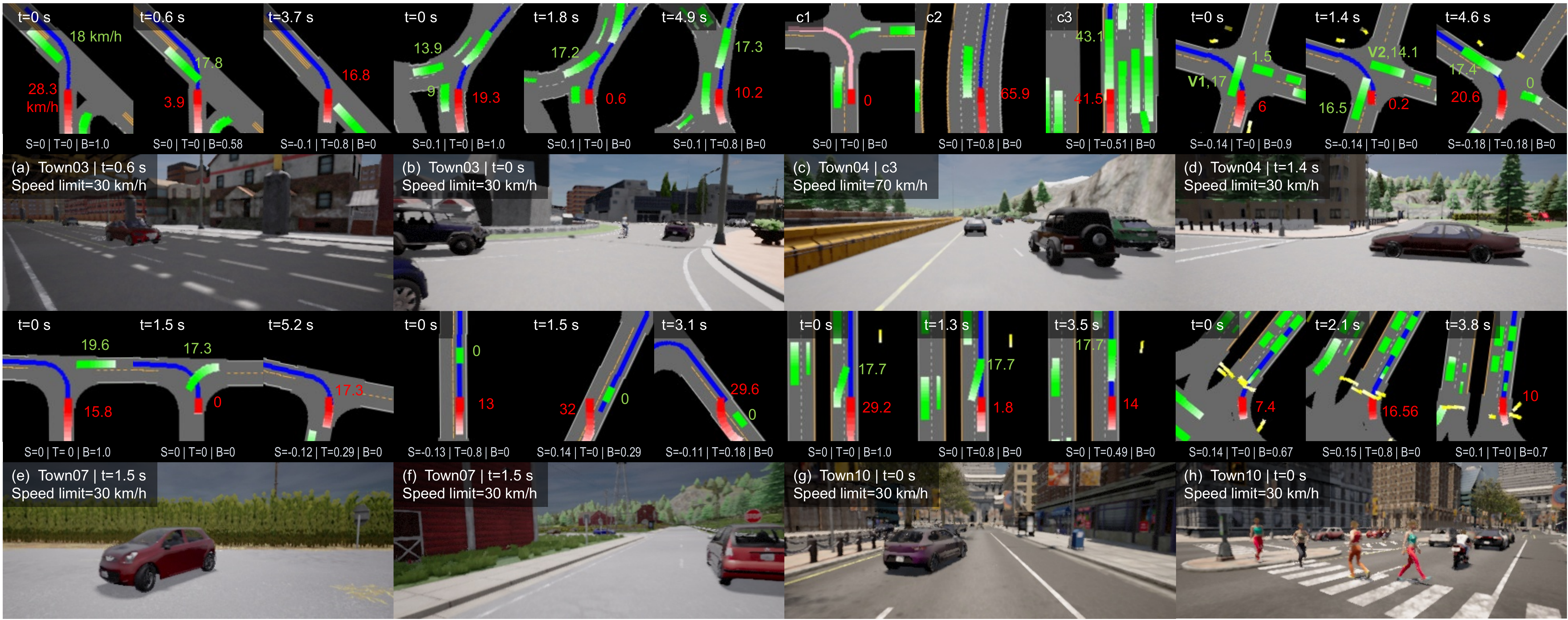}
        \caption{Closed-loop evaluation results of our \texttt{DiGNet}. We show several driving clips with BEV and FPV images in four maps covering urban, rural and highway areas, where Town04 and Town10 are unseen maps during training. For better visualization, we set the color of route to pink if traffic light turns to red, otherwise to blue. In addition, we render trajectories in the past 1.5 s for road agents, where the lighter color indicates the more distant historical location. We also label the speed of key vehicles (ego-vehicle in red and other vehicles in green), and the output control commands (S-steer, T-throttle, B-brake) for better understanding. The range of steering is [-1,1], while for throttle and brake the range is [0,1]. The sample driving behaviors are: slowing down when taking unprotected turns at (a,d,e) intersections or (b) roundabouts for collision avoidance, (c1) stopping at intersections with red traffic light, (c2,c3) high-speed driving and vehicle-following on highways, (f) deviating the route a bit to bypass a parked vehicle on narrow roads, (g) timely slowing down when an aggressive vehicle in front suddenly changes lanes to the lane of ego-vehicle, and (h) crowd-aware safe driving behaviors among multiple pedestrians, including slowing down for collision avoidance (t=0 s) and continuing to drive when no pedestrians blocking ways (t=2.1 s). }
        \label{fig:qualitative}
        \vspace{-0.35cm}
\end{figure*}

\subsection{Evaluation}
\label{subsec:evaluation}
\subsubsection{Benchmark} In order to cover as many driving scenarios as possible to thoroughly evaluate different methods, we choose the most complicated six maps in CARLA, including two unseen maps \texttt{Town04} and \texttt{Town10} for closed-loop driving tasks. \texttt{Town04} is a mixed environment with highway roads and a small town. More concretely, in each map, we set 10 routes for the ego-vehicle to drive, and each route corresponds to 40 driving episodes. For each episode, we set random number of vehicles ($10\sim300$) and pedestrians ($10\sim90$) navigating around the map. They are also spawned with random properties (i.e., speeds, destinations) at random positions. Due to such randomness, the environmental dynamics of seen maps is also different from the training dataset. Finally, we conduct 16,800 episodes to evaluate seven driving models, covering a distance of 7,102 km. Note that compared with previous benchmark methods introduced in \cite{dosovitskiy2017carla, codevilla2019exploring}, ours covers much more traffic scenarios, and is more suitable to evaluate the self-driving performance.

\subsubsection{Metrics} We use two metrics to measure the driving performance on each map. The first is success rate (SR). An episode is considered to be successful if the agent reaches a certain goal without any collision. The episode will be recounted if the agent gets stuck in traffic jam. The second metric is driving score (DS) defined as $\frac{1}{n}\sum_i^nR_iP_i$, where $n$ stands for the number of episodes for each town (400 in this work), $R_i$ is the percentage of completion of the route in the i-th episode, and $P_i$ is the collision penalty of the i-th episode (set to 0.5 if collision happens, otherwise set to 1).

\subsubsection{Quantitative Analysis} Tab. \ref{tab:evaluation} shows the quantitative evaluation results. For clarity, in the following we list our main findings and corresponding interpretations:

- Semantic scene representations are more suitable than raw sensory data to learn \textit{generalizable} self-driving policies. We can see that the performance drop from seen maps to unseen maps is quite distinct for \texttt{CILRS}. For example, when the testing environment changes from \texttt{Town03} to \texttt{Town10}, the SR of \texttt{CILRS} significantly drops by 40 times (43.2\% $\rightarrow$ 1.2\%), while the performance of state-based methods does not degrade that much.

- GNNs are more suitable than MLP to handle interactions among different nodes. As we can see, $\texttt{MLP}$ performs worst among the state-based methods. For example, the SR in \texttt{Town10} of \texttt{MLP} is only 38.2\%, which is much lower than the GNN-based methods ($55.2\%\sim68.0\%$).



- Properly combining both the context and state information can help improve the driving ability, as our method \texttt{DiGNet} achieves the best overall results. We conjecture that the pixel information, which is encoded as the latent vector $\bm{z}$ in this work, can provide complementary spatial features (e.g., road structures and lane positions) on the basis of graph states.

- Using mid-level control outputs ($\kappa_v,\kappa_{\theta}$) rather than direct command signals can achieve better driving results, as \texttt{DiGNet} achieves higher SR and DS than \texttt{DiGNet(CTL)} in all maps.

\subsubsection{Qualitative results}
Fig. \ref{fig:qualitative} shows the qualitative results of \texttt{DiGNet}. We can see that our model can safely drive in diverse dynamic environments with different road structures (roundabouts, intersections, highways, etc.) and traffic scenarios, whilist obeying traffic rules such as speed limits and traffic lights. For example, in Fig. \ref{fig:qualitative}-(d), our model is turning left at an intersection, however, vehicle V1 is coming from the opposite direction without slowing down (speed=17 km/h). Our model timely looses the throttle and applies a large brake (0.9) to avoid collisions. When t=1.4 s, another vehicle V2 is coming from the right side at a speed of 14.1 km/h. To avoid collision, our model continues to wait until it is safe to speed up (t=4.6 s).

\begin{figure}[t]
        \centering
        \includegraphics[width = 0.85\columnwidth]{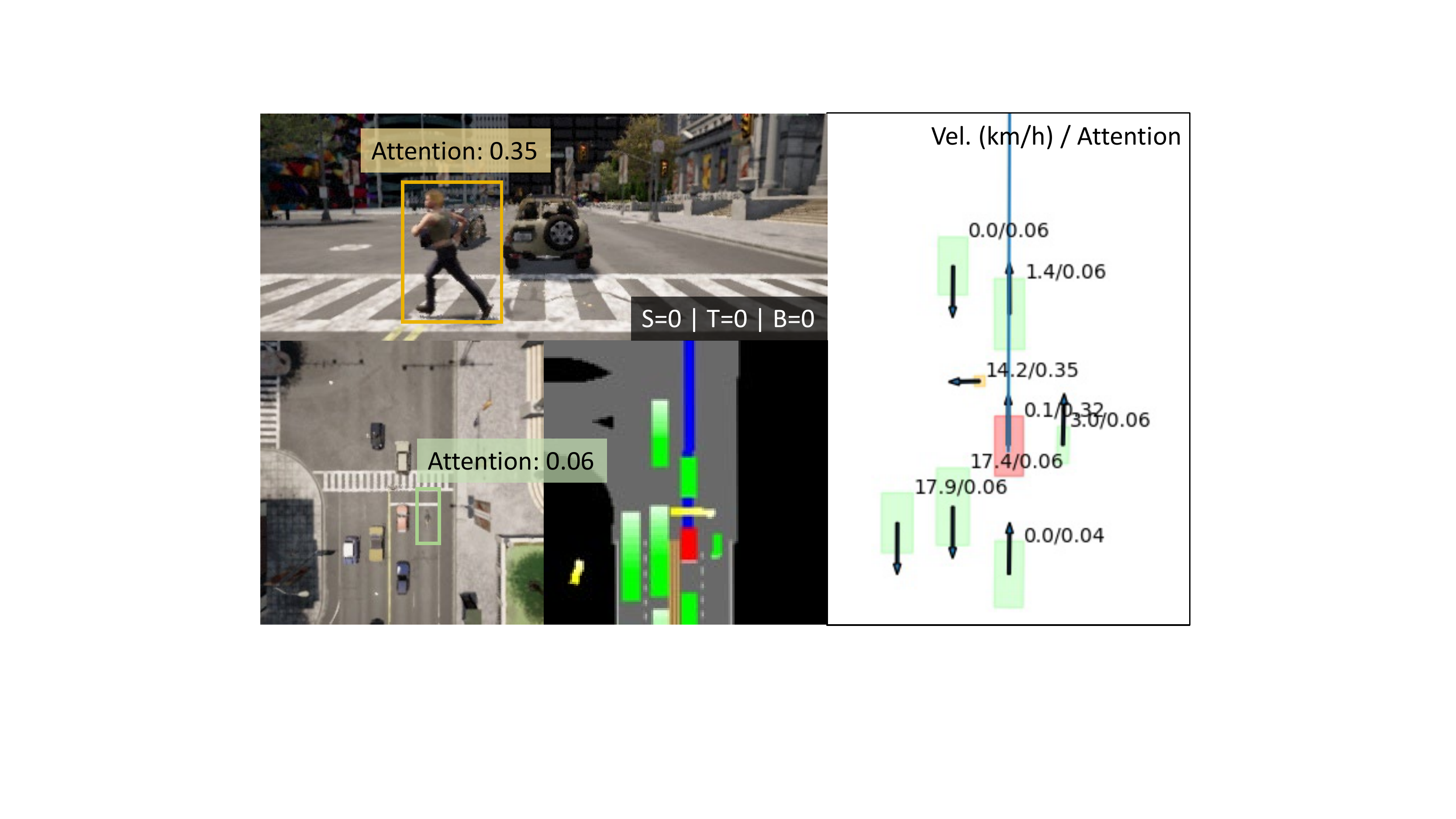}
        \caption{A sample driving scenario where the ego-vehicle stops for pedestrians with our $\texttt{DiGNet}$ (left). We also show the corresponding attentions for the road agents from the GAT module (right).}
        \label{fig:attention}
\end{figure}

\subsubsection{Attention Analysis}
To show the benefits of GAT against other GNN methods in our task, we analyze its attention outputs using Fig. \ref{fig:attention}. We can see that there is a pedestrian (bounded by the orange box) running across the street in front of the ego-vehicle, which exerts a strong influence with higher attention (0.35) than the other agents ($0.04\sim0.06$), therefore, the ego-vehicle stops at the intersection to avoid an accident by applying a zero throttle value. Note that there is a motorcycle (bounded by green box) at the right lane, and it is also very close to the ego-vehicle but assigned with a lower attention (0.06), because it does not have much influence on the ego-vehicle. This phenomenon demonstrates that the ideas of \texttt{GCN(U)} and \texttt{GCN(D)} are not reasonable in some occasions, because the influences of road agents are not always equal or related with distance. By contrast, \texttt{GAT} is more suitable in our task. It dynamically assigns attentions using a self-attention mechanism based on specific driving contexts. Such difference also explains why GCN methods generally produce worse results than \texttt{GAT} in Tab. \ref{tab:evaluation}.




\section{Conclusion and Future Work}
In this work we developed a context-aware graph-based deep navigation network named DiGNet to achieve scalable self-driving in generic traffic scenarios, such as unprotected left turns, narrow roads, roundabouts, pedestrian- and vehicle-rich intersections. More specifically, we first used VAE to encode semantic driving contexts into concise and informative latent vectors, which can then be incorporated with the state information (e.g., locations and speeds) into GNNs to model the complex interactions among road agents. The large-scale closed-loop evaluation results revealed that our method achieves the highest success rates and driving scores in most environments. However, we mainly adopted the supervised imitation learning method to train driving policies, but this method may suffer from the distribution mismatch and dataset bias problems. In the future, we will investigate how to alleviate these issues using deep reinforcement learning techniques for better self-driving performance.

\bibliographystyle{IEEEtran}
\bibliography{main.bib}

\end{document}